\documentclass[10pt,twocolumn,letterpaper]{article}

\usepackage{cvpr}              

\nolinenumbers
\definecolor{cvprblue}{rgb}{0.21,0.49,0.74}
\usepackage[pagebackref,breaklinks,colorlinks,citecolor=cvprblue]{hyperref}
\usepackage{amssymb}
\def\paperID{*****} 
\def\confName{CVPR}
\def\confYear{2024}

\title{Multimodal Fusion Method with Spatiotemporal Sequences and Relationship Learning for Valence-Arousal Estimation}

\author{
Jun Yu$^1$, Gongpeng Zhao$^1$, Yongqi Wang$^1$\thanks{Corresponding author} , Zhihong Wei$^1$\thanks{Corresponding author} , Zerui Zhang$^1$, Zhongpeng Cai$^1$,\\ 
Guochen Xie$^1$, Jichao Zhu$^1$, Wangyuan Zhu$^1$, Yang Zheng$^1$\\
$^1$University of Science and Technology of China\\
\tt\small harryjun@ustc.edu.cn\\
\tt\small \{zgp0531,wangyognqi,weizh588,igodrr,zpcai,\\ \tt\small xiegc,jichaozhu,zhuwangyuan,zhengyang\}@mail.ustc.edu.cn \\
}

\begin{document}

\maketitle

\begin{abstract}
This paper presents our approach for the VA (Valence-Arousal) estimation task in the ABAW6 competition. We devised a comprehensive model by preprocessing video frames and audio segments to extract visual and audio features. Through the utilization of Temporal Convolutional Network (TCN) module, we effectively captured the temporal and spatial correlations between these features. Subsequently, we employed a Transformer encoder structure to learn long-range dependencies, thereby enhancing the model's performance and generalization ability. Additionally, we proposed the LA-SE module to better capture local image information and enhance channel selection and suppression. Our method leverages a multimodal data fusion approach, integrating pre-trained audio and video backbones for feature extraction, followed by TCN-based spatiotemporal encoding and Transformer-based temporal information capture. Experimental results demonstrate the effectiveness of our approach, achieving competitive performance in VA estimation on the AffWild2 dataset.
\end{abstract}    
\section{Introduction}
\label{sec:intro}

With the continuous development of artificial intelligence technology and society, human-computer interaction has become a research field of great concern. An excellent human-computer interaction system not only needs to have efficient functionality but also needs to consider the user's emotions and experiences. Emotion analysis plays an important role in this regard, as it is expected to provide a more accurate understanding of human emotions, thereby designing human-computer interaction systems that are more user-friendly and closer to user needs.

In recent years, human emotional behavior analysis has received increasing attention. Commonly used human expression modalities\cite{kollias2020face,Kollias_2019,kollias2023multi} in this field include Action Units (AU)\cite{kollias2021distribution} basic expression categories (EXPR)\cite{kollias2021analysing}, and Valence-Arousal (VA)\cite{kollias2021affect}. Action Units describe the movement of specific facial regions and are the smallest units for describing expressions. Basic expression categories classify expressions into several emotional categories such as happiness, sadness, etc. VA is a model that includes two continuous values representing the valence and arousal of emotions, which can better describe human emotional states.
The ABAW workshop\cite{kollias20246th,kollias2023abaw2,kollias2022abaw} aims to address the challenges of in-the-wild emotional behavior analysis, which reflects a key characteristic of real-world Human-Computer Interaction (HCI) systems. The objective is to develop intelligent machines and robots capable of understanding human emotions, moods, and behaviors. By interacting with humans in a "human-centered" manner, these systems provide engaging experiences and effectively serve as digital assistants. Such interactions should not be constrained by individual backgrounds, age, gender, race, education level, occupation, or social status. Therefore, the development of intelligent systems capable of accurately analyzing human behavior in the wild is essential for fostering trust, understanding, and intimacy between humans and machines in real-life environments. The advancement of this technology holds promise for the future of HCI, enabling more sophisticated and human-centric experiences in everyday life. To promote the development of human emotional behavior analysis, the Affective Behavior Analysis in-the-wild (ABAW) competition has proposed a series of challenges aimed at overcoming obstacles in this field.  The ABAW competition has constructed large-scale multimodal video datasets, such as Affwild \cite{Zafeiriou2017AffWildVA, Kollias_2019,kollias2019deep,kollias2019face}and Affwild2\cite{kollias2019expression, kollias2021affect,kollias2022abaw,kollias2020analysing}, which provide valuable resources for researchers, greatly advancing the progress of in-the-wild facial expression analysis and accelerating the application in related industries. The Affwild2 dataset contains a large number of videos, most of which are annotated frame by frame with AU, basic expression categories, VA, and other labels, providing researchers with rich annotated data.

Emotion recognition\cite{usman2018using, KHARE2024102019}, as a critical task, plays a crucial role in human emotion understanding and human-computer interaction. Among them, the estimation of Valence and Arousal is a key component of emotion recognition. Through the dimensional model\cite{Sandbach2012StaticAD}, we can view emotional states as points in continuous space, with Valence and Arousal serving as axes. However, achieving accurate VA estimation faces numerous challenges. Firstly, the expression of emotions is subjective and varies between individuals, leading to uncertainty in emotion recognition results due to differences in how different individuals perceive emotions. Secondly, emotional expression is typically dynamic and influenced by time, necessitating consideration of long-term temporal dependencies to better capture the process of emotional change. Additionally, emotional expression takes various forms and may be influenced by external factors and individual experiences, requiring emotion recognition models to have a certain level of robustness to adapt to different contexts and individual characteristics. Addressing these challenges requires integrating multimodal data and advanced models to improve the accuracy and robustness of VA estimation, thereby advancing the application and development of emotion recognition technology in the fields of human-computer interaction and affective computing.

To address the challenges above, we employ a multimodal data fusion approach. Firstly, we utilize a pre-trained audio VGGnet as the backbone for extracting dynamic audio features from VGGish\cite{hershey2017cnn}, and an IResnet\cite{duta2020improved} pre-trained with the arcface\cite{Deng_2022} method as the backbone for extracting video frames. Consequently, the expressiveness of deep features in both visual and audio modalities can be further enhanced through fine-tuning. Subsequently, for each branch, we utilize CNN backbones to extract dynamic spatial depth features from video frames and log-spectrograms, followed by the use of Temporal Convolutional Networks\cite{bai2018empirical} to further learn spatiotemporal encoding. Additionally, we integrate the LA-SE module before the TCN module to better capture local image information and enhance channel selection and suppression. The LA-SE module combines LANet for spatial aggregation and SENet for channel-wise selection, improving the effectiveness of feature extraction. Lastly, these features are concatenated and fed into a Transformer structure to capture temporal information for downstream tasks. During training, we employ a strategy of large-window resampling, where the window size determines the amount of contextual information the model considers for predicting each time step.

The main contribution of the proposed method can be summarized as:

1. We introduce a novel approach for fusing multimodal data, leveraging pre-trained audio and video backbones to extract dynamic features from both visual and audio modalities.

2. To enhance feature extraction and selection, we propose the LA-SE network, combining LANet for spatial aggregation and SENet for channel-wise selection. This module captures local image information effectively, improving model performance.

3. Utilizing Temporal Convolutional Networks (TCN), we further learn spatiotemporal encoding to capture temporal dependencies and patterns in the data.

4. We utilize a Transformer structure to effectively capture temporal information for downstream tasks, ensuring robust performance in real-world scenarios.

\section{Related Work}
\label{sec:formatting}

In the field of affective behavior analysis, multimodal feature fusion is a key issue. Most studies\cite{ortega2019emotion,tzirakis2017end} adopt multimodal methods to improve model performance by fusing visual and audio features. In previous research, researchers have proposed various fusion methods, such as directly fusing multimodal features into a common feature vector for analysis\cite{nagrani2021attention}, or separately analyzing the features of each modality and then fusing the final outputs to obtain better prediction results\cite{zhang2022transformer}, or even combining the above two fusion methods to achieve better results\cite{meng2022multi,kuhnke2020two}combined visual and audio information in videos, constructing a dual-stream network for emotion recognition, achieving high performance. Currently, convolution-based methods\cite{poria2016convolutional,zhang2020m} have made significant progress. At the same time, the application of Transformer in multimodal learning has become mainstream, \cite{parthasarathy2021detecting,han2022survey,kim2022facial}achieving excellent performance using the Transformer structure. In terms of audio features\cite{zhang2018attention,stuhlsatz2011deep,lieskovska2021review}, energy features, temporal features, etc., are widely used to improve the ability of emotion recognition.

Recently, researchers have proposed many new methods and frameworks\cite{deng2021iterative,vu2021multitask,wang2021multi,zhang2021prior}, including tasks such as AU detection, expression recognition, and VA estimation. Some of these methods leverage the correlation between VA and AU or VA and EXPR, proposing multi-task frameworks. These methods can extract complementary information from other tasks. To address the high cost of annotating emotion/AU/VA labels from real-world facial images, some researchers have proposed using self-supervised learning (SSL) methods to leverage knowledge from existing large-scale unlabeled data. These methods provide new avenues for improving the accuracy of affective analysis.

In conclusion, multimodal features play an important role in affective behavior analysis, and various fusion methods and self-supervised learning methods provide new ideas and approaches for improving the accuracy and efficiency of emotion recognition.
\begin{figure*}[ht]
\centering
\includegraphics[width=\linewidth]{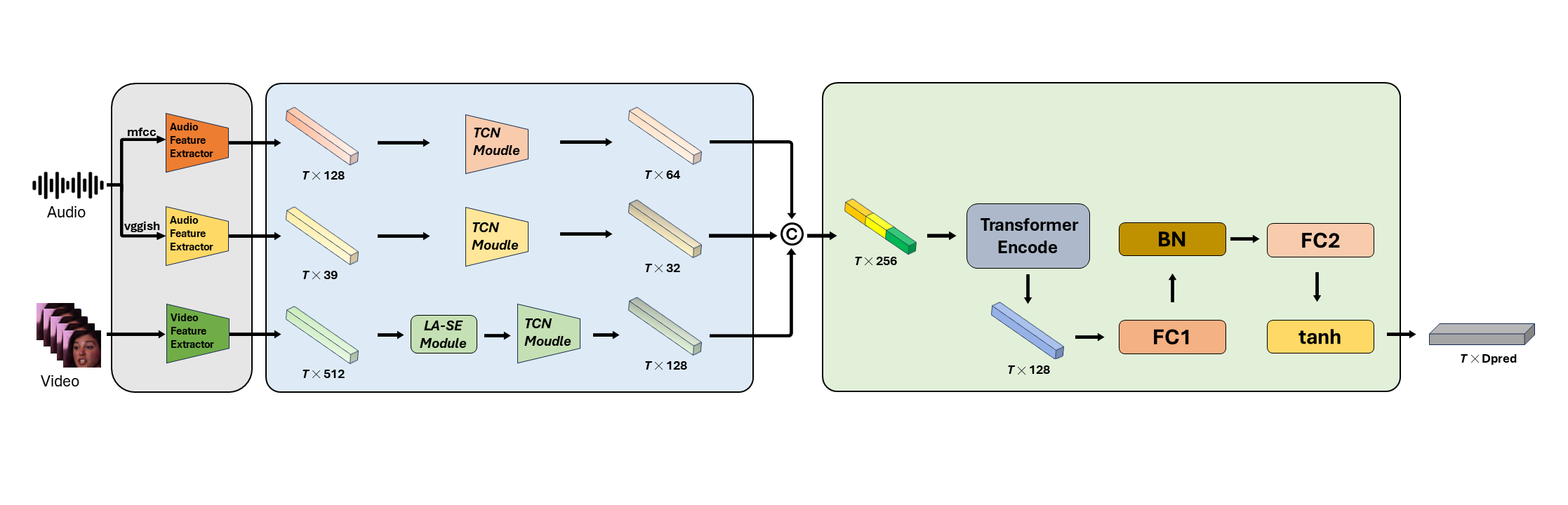}
\caption{Our proposed framework for VA estimation. The model begins with preprocessing of video frames and audio segments, followed by feature extraction using pre-trained audio and video backbones. The LA-SE module enhances local image information capture and channel selection. Temporal Convolutional Network (TCN) modules capture temporal and spatial correlations in the features, while a Transformer encoder structure learns long-range dependencies. }
\label{Figure 1}
\end{figure*}
\section{Proposed approach}
In this section, we focus on introducing our method. To better accomplish the VA estimation task, we have designed an efficient network. As shown in Figure \textcolor{red}{1}, we use a CNN backbone to extract dynamic spatial depth features from video frames and audio. Subsequently, we utilize TCN to further learn spatiotemporal encoding. Afterward, we concatenate the video and audio features and feed the output into a fully connected layer to apply to the VA estimation task.
\subsection{Feature Extraction}
\textbf{VGGish:}VGGish\cite{hershey2017cnn} is a pre-trained deep neural network model developed by Google, specifically designed for audio feature extraction. It is commonly used for tasks such as audio classification, audio event detection, and other audio-related tasks. The model architecture is based on the VGG network architecture, which consists of multiple convolutional layers followed by max-pooling layers. Its output is a 128-dimensional feature vector that can be used for speech-related tasks.

\textbf{MFCC:}MFCC\cite{davis1980comparison} is a widely used method for audio feature extraction in tasks such as speech recognition and audio processing. It was proposed by American engineer Harvey Fletcher and Japanese engineer Takeo Yamazaki in the 1960s. Based on the perceptual characteristics of the human ear to sound frequencies, MFCC transforms audio signals into a set of feature vectors to represent the spectral characteristics of the audio. The process of extracting MFCC involves segmenting the audio signal into frames, applying windowing, performing a fast Fourier transform to compute the power spectrum, applying a Mel filterbank, taking the logarithm, and finally applying discrete cosine transform to obtain the MFCC coefficients. These coefficients can be used to train machine learning models, such as speech recognition and speaker recognition, for the analysis and processing of audio data. Its output is a 39-dimensional feature vector that can be used for speech-related tasks.

\textbf{IResNet-50:}IResNet-50\cite{behrmann2019invertible} is a variant of the ResNet\cite{he2015deep} architecture. IResNet-50 incorporates the idea of residual connections to enable the training of very deep neural networks more effectively.  In this study, we trained IResNet-50 using the AffectNet\cite{Mollahosseini_2019} dataset and obtained a 512-dimensional visual feature vector.

\begin{table}[htbp]
\centering
\caption{The dimensions of features.}
\label{tab:feature_dimensions}
\begin{tabular}{ccc}
\toprule
\textbf{Feature Modality} & \textbf{Dimension} & \textbf{Description} \\ \midrule
VGGish & 128 & A \\
MFCC & 39 & A \\
IResNet-50 & 512 & V \\ \bottomrule
\end{tabular}
\end{table}

\subsection{Preprocessing}

Our preprocessing approach is primarily based on the solution proposed by Su Zhang, whoes were last year's third-place winners\cite{zhang2023multimodal}. In our experiments, the input consists of one visual feature and two audio features, but their sizes are typically different, and there may even be significant discrepancies. The sizes of the features are shown in Table \textcolor{red}{1}. The visual preprocessing first resizes all images to a size of 48×48×3, based on the cropped-aligned image data provided by the organizer. For each trial in the training or validation set, its length is determined by the number of rows in the annotation text file, excluding those marked as -5. In the test set, the length is determined by the frame count of the original video. A zero matrix of size N×48×48×3 is initialized, and then each row is assigned to the corresponding jpg image if it exists. The audio preprocessing converts all videos to mono with a 16K sampling rate in WAV format initially. Then, the logmelspectrogram is extracted using preprocessing code from the Vggish repository, with the hop length specified to synchronize with other modalities and annotations. Emotion label processing involves excluding rows containing -5. To ensure consistency in length between features and annotations, the feature matrix is either padded repeatedly (using the last feature points) or trimmed (from the rear), depending on whether the feature length is shorter or longer than the trial length.

\subsection{Training Details}
In this study, we employed a six-fold cross-validation method to fully utilize the AffWild2 database. The database consists of 360 training trials, 72 validation trials, and 162 testing trials, covering various emotional expressions and contexts. Through cross-validation, we were able to leverage the information in the dataset effectively, reducing the model's dependence on specific data distributions, thereby better assessing the performance and robustness of the model. Given the video frames $V_i$ and the corresponding audio segments $A_i$ obtained through preprocessing, we employ pre-trained video feature extraction models (IResNet-50) and audio feature extraction models (VGGish) to extract visual features $F_{\text{vis}}^i$ and audio features $F_{\text{aud}}^i$ for each frame separately. Subsequently, we send them to TCN modules for processing to capture the temporal and spatial correlations between them. As shown in Equation \textcolor{red}{1}, we concatenate the extracted visual frame feature branch and the audio feature branch. Next, the processed features are fed into a Transformer encoder structure, which helps learn long-range dependencies between features and improves the performance and generalization ability of the model. The Transformer encoder consists of four encoder layers, each with a dropout rate of 0.4. The output of the encoder is then passed through a fully connected layer, a Batch Normalization layer, and another fully connected layer. Finally, a tanh activation function is used for prediction, yielding the Valence (V) or Arousal (A) values for each video frame. This design enhances the model's expressive power and stability, enabling it to better adapt to various tasks and data distributions. When computing the loss, we utilize the CCCLoss as shown in Equation \textcolor{red}{2}.Due to the frame-wise prediction process, we adopt a smoothing strategy to enhance the stability of the model. We observed occasional omissions of frames in the cropped face images extracted from the videos. Since video frames are continuous, we replace the missing frames with the last available frame.
\begin{equation}
f = \text{concat}(\hat{f}_1, \hat{f}_2, \hat{f}_3)
\end{equation}
\begin{equation}
\text{CCCLoss} = 1 - CCC(\hat{v}_{\text{frame}_i}, v_{\text{frame}_i}) + 1 - CCC(\hat{a}_{\text{frame}_i}, a_{\text{frame}_i})
\end{equation}

\subsection{LA-SE Module}
To better capture local image information and enhance channel selection and suppression, we propose the LA-SE network, which combines the LANet\cite{yang2022dynamic} and SENet\cite{hu2019squeezeandexcitation}. In Figure \textcolor{red}{2}, LANet is employed to aggregate local information, accomplished through a sequence of two consecutive $1 \times 1$ convolutional layers. These layers consolidate spatial information into a unified channel, which is then subjected to scaling using a sigmoid function to produce spatial attention.In contrast, Figure  \textcolor{red}{3} illustrates the utilization of SENet for channel-wise selection and suppression. SENet comprises two primary operations: squeeze and excitation.

The squeeze operation compresses global channel information into a single-channel descriptor using global average pooling:
\begin{equation}
z_t = \frac{1}{w \times h} \sum_{i=1}^{w} \sum_{j=1}^{h} u_t(i, j)
\end{equation}

where $z_t$ represents the signal of channel $t$ and $u_t(i, j)$ denotes the element of channel $t$ at position $(i, j)$. The excitation operation models channel-wise dependencies through two fully connected (FC) layers:
\begin{equation}
s = \sigma \left( w_2 \cdot g(w_1 \cdot z) \right)
\end{equation}
where $\sigma$ refers to the sigmoid function, $g$ is the ReLU function, and $w_1$ and $w_2$ are the weights of the FC layers.

Finally, the scaling operation adjusts each channel dynamically based on learned activations:
\begin{equation}
x_i = s_i \times u_i
\end{equation}
where $s_i$ represents a scalar related to channel $i$, and $u_i$ is the input for channel $i$.

The LA-SE network integrates the advantages of LANet and SENet to extract more representative and robust features for image understanding tasks, thereby improving model performance and generalization capability.

\subsection{TCN Module}

The Temporal Convolutional Network (TCN) module is utilized to capture temporal dependencies between frames and features. It consists of a stack of dilated convolutional layers followed by activation functions and pooling operations. The dilated convolutional layers have exponentially increasing dilation rates, enabling them to capture information from a wider range of temporal contexts.

Formally, given an input sequence $X = \{x_1, x_2, ..., x_T\}$, where $x_t$ represents the feature representation at time step $t$, the output of the TCN module can be calculated as follows:

\begin{equation}
y_t = \text{ReLU}\left(\sum_{i=1}^{N} W_i * x_{t + (i-1)d}\right)
\end{equation}

where $*$ denotes the convolution operation, $W_i$ represents the learnable weights of the $i$-th convolutional layer, $N$ is the number of layers, and $d$ is the dilation rate. The ReLU function introduces non-linearity to the model, allowing it to learn complex temporal patterns.

The output of the TCN module, denoted as $Y = \{y_1, y_2, ..., y_T\}$, contains features that capture temporal dependencies and can be further processed by subsequent layers for downstream tasks such as classification or regression.

\subsection{Transformer Encode}
The Transformer Encoder is a crucial component in many sequence modeling tasks, including natural language processing and video analysis. It utilizes self-attention mechanisms to capture long-range dependencies within sequences effectively.

Formally, given an input sequence $\mathbf{X} = \{\mathbf{x}_1, \mathbf{x}_2, ..., \mathbf{x}_T\}$, where $\mathbf{x}_t$ represents the feature representation at time step $t$, the Transformer Encoder processes the sequence through multiple layers of self-attention and feed-forward neural networks. Each layer consists of two sub-layers: a multi-head self-attention mechanism and a position-wise fully connected feed-forward network.

The multi-head self-attention mechanism computes attention scores between all pairs of positions in the input sequence, allowing the model to focus on relevant information from distant positions. This is achieved by linearly projecting the input into multiple lower-dimensional representations, computing attention scores, and then aggregating them across different heads.

The position-wise feed-forward network applies a two-layer fully connected neural network independently to each position in the sequence. This operation introduces non-linearity and enables the model to capture complex interactions between features.

After processing through multiple layers of self-attention and feed-forward networks, the Transformer Encoder produces a sequence of output representations $\mathbf{Y} = \{\mathbf{y}_1, \mathbf{y}_2, ..., \mathbf{y}_T\}$, which captures rich contextual information and dependencies within the input sequence. These representations can be further used for downstream tasks such as classification, regression, or sequence generation.

\begin{figure}[ht]
    \centering
    \begin{minipage}{\linewidth}
        \centering
        \includegraphics[width=\linewidth]{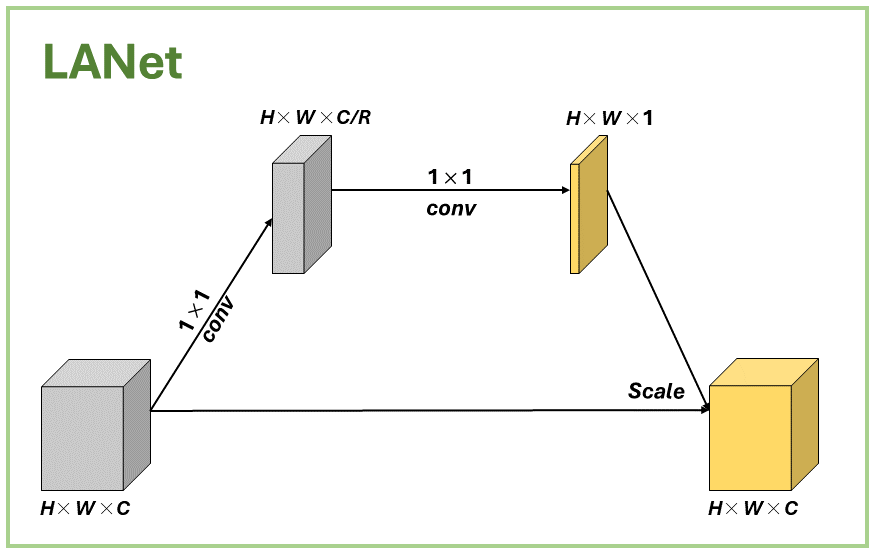}
        \caption{The LANet module, where H, W and C refer to height, width and number of channels, respectively.}
        \label{fig:lanet}
    \end{minipage}
    \begin{minipage}{\linewidth}
        \centering
        \includegraphics[width=\linewidth]{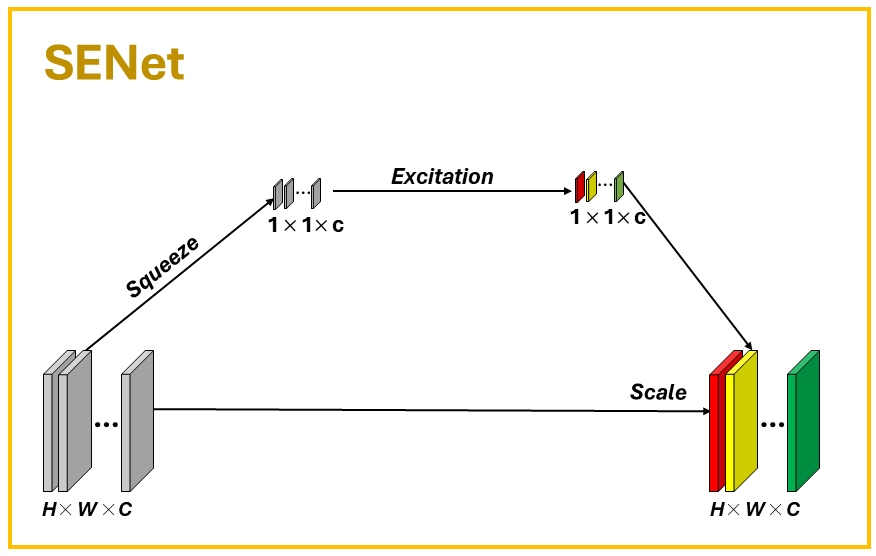}
        \caption{The SENet module, where H, W and C refer to height, width and number of channels, respectively.}
        \label{fig:senet}
    \end{minipage}
\end{figure}

\begin{table*}[htbp]
    \centering
    \caption{Module Ablation Experiment Results for LA-SE, TCN, and Transformer Encode}
    \label{tab:module_ablation}
    \begin{tabular}{l c c c c}
        \toprule
        \textbf{Experimental Combination} & \textbf{LA-SE} & \textbf{TCN} & \textbf{Transformer Encode} & \textbf{CCC Score (\%)} \\
        \midrule
        Baseline &  &  &  & 31.08 \\
        +LA-SE & \checkmark &  &  & 33.96 \\
        +TCN &  & \checkmark &  & 51.57 \\
        +Transformer Encode &  &  & \checkmark & 38.77 \\
        +LA-SE + TCN & \checkmark & \checkmark &  & 52.71 \\
        +LA-SE + Transformer Encode & \checkmark &  & \checkmark & 40.12 \\
        +TCN + Transformer Encode &  & \checkmark & \checkmark & 60.21 \\
        +LA-SE + TCN + Transformer Encode & \checkmark & \checkmark & \checkmark & 63.56 \\
        \bottomrule
    \end{tabular}
\end{table*}

\section{Experiment}
In this section, we provided an overview of the dataset, evaluation protocol, and experimental results used in the study.
\subsection{Dataset and Evaluation}
\textbf{Affwild2 dataset.} Aff-Wild2 is an extension of the Aff-Wild dataset for affect recognition. It approximately doubles the number of included video frames and the number of subjects; thus, improving the variability of the included behaviors and of the involved persons. Aff-Wild2 is a significant research asset, being the only database annotated for all three main behavioral tasks in the wild. It is a large-scale database and the first one to feature AU annotations alongside both audio and video. Annotations in Aff-Wild2 are frame-based, encompassing seven basic expressions, twelve action units, as well as valence and arousal. In total, Aff-Wild2 comprises 564 videos, around 2.8 million frames, involving 554 subjects. These subjects represent diverse demographics in terms of age, ethnicity, and nationality, while also presenting a wide array of environmental and situational variations.

\textbf{Evaluation.}To estimate the VA, we computed the Concordance Correlation Coefficient (CCC) separately for arousal and valence. The evaluation metric for this competition is represented by Equation \textcolor{red}{7}.

\begin{equation}
P = 0.5 \times (CCC_{\text{arousal}} + CCC_{\text{valence}})
\end{equation}

\subsection{Module Ablation Experiment Results}

In this section, we present the results of the module ablation experiments for LA-SE, TCN, and Transformer Encode, along with their impact on the CCC (Concordance Correlation Coefficient) scores.

The table \textcolor{red}{2} presents the CCC scores (\%) obtained from different experimental combinations. Each row represents a specific experimental configuration, including individual modules and their combinations. Starting with the Baseline model, which directly concatenates features extracted from video and audio, yielding a CCC score of 31.08. Subsequently, we gradually added different modules to investigate their effects on model performance. By adding the LA-SE module, the CCC score increased to 33.96, indicating an improvement in model performance. When the TCN module was added alone, the CCC score significantly increased to 51.57, demonstrating its effectiveness in capturing temporal dependencies between frames. The introduction of the Transformer Encode module further improved performance, increasing the CCC score to 38.77. This suggests that Transformer Encode helps to learn long-range dependencies between features, enhancing the model's generalization capability. When combining the LA-SE and TCN modules, the CCC score further improved to 52.71, demonstrating the synergistic effect of these two modules. Similarly, the combination of LA-SE and Transformer Encode also led to a performance improvement, with a CCC score of 40.12. However, the most significant improvement occurred with the combination of TCN and Transformer Encode, resulting in a CCC score of 60.21. This indicates that TCN and Transformer Encode modules complement each other well in capturing temporal and feature relationships. Finally, when all three modules were combined, the model achieved the highest performance, with a CCC score of 63.71. This further validates the importance and effectiveness of the LA-SE, TCN, and Transformer Encode modules in improving model performance.

\subsection{Results on Validation set}
For the estimation of VA, we evaluate the models based on the average CCC values for valence and arousal. To enhance the model's generalization, we further conduct a six-fold cross-validation on randomly segmented annotated data. Detailed experimental results are provided in Table \textcolor{red}{3}. This table presents the Valence and Arousal scores for different validation sets (fold-0 to fold-5). Each validation set corresponds to a specific cross-validation fold. Each row represents a validation set, including its Valence and Arousal scores. The last row shows the best Valence and Arousal scores obtained across all folds, highlighted in bold.

\begin{table}[htbp]
\centering
\caption{Valence and Arousal Scores for Different Validation Sets}
\label{tab:validation_scores}
\begin{tabular}{@{}ccc@{}}
\toprule
\textbf{Val Set} & \textbf{Valence} & \textbf{Arousal} \\
\midrule
fold-0 & 0.5121 & 0.5507 \\
fold-1 & 0.5789 & 0.5992 \\
fold-2 & 0.5547 & 0.5862 \\
fold-3 & 0.5743 & 0.5959 \\
fold-4 & 0.5971 & 0.6156 \\
\textbf{fold-5} & \textbf{0.6123} & \textbf{0.6589} \\
\bottomrule
\end{tabular}
\end{table}

\section{Conclusion}
This paper introduces our submission for the VA estimation task in ABAW6 competition. We conducted preprocessing on video frames and audio segments to extract visual and audio features, constructing a comprehensive model for VA (Valence-Arousal) estimation task. Utilizing TCN modules, we captured the temporal and spatial correlations between features. Finally, we fed the extracted features into a Transformer encoder structure to learn long-range dependencies and enhance the model's performance and generalization ability.
{
    \small
    \bibliographystyle{ieeenat_fullname}
    \bibliography{main}
}


\end{document}